Natural language processing to automatically extract the presence and severity of esophagitis in notes of patients undergoing radiotherapy

Shan Chen,[1,2] Marco Guevara,[1,2] Nicolas Ramirez,[1,2] Arpi Murray,[2] Jeremy L. Warner,[3,4] Hugo JWL Aerts,[1,2,5] Timothy A. Miller,[6] Guergana K. Savova,[6] Raymond H. Mak,[1,2] Danielle S. Bitterman,[1,2]
1. Artificial Intelligence in Medicine (AIM) Program, Mass General Brigham, Harvard Medical School, Boston, MA
2. Department of Radiation Oncology, Brigham and Women's Hospital/Dana-Farber Cancer Institute, Boston MA
3. Population Sciences Program, Legorreta Cancer Center, Brown University, Providence, RI, U.S.A.
4. Lifespan Cancer Institute, Providence, RI, U.S.A.
5. Radiology and Nuclear Medicine, GROW & CARIM, Maastricht University, The Netherlands
6. Computational Health Informatics Program, Boston Children's Hospital, Boston, MA

Corresponding author:
Dr. Danielle S. Bitterman
Department of Radiation Oncology
Dana-Farber Cancer Institute/Brigham and Women's Hospital
75 Francis Street, Boston, MA 02115
Email: Danielle_Bitterman@dfci.harvard.edu
Phone: (857) 215-1489
Fax: (617) 975-0985

Running head: NLP for esophagitis severity extraction.

Funding statement: DSB: Woods Foundation, Jay Harris Junior Faculty Award, Joint Center for Radiation Therapy Foundation.
Disclosures: DSB: Associate Editor of Radiation Oncology, HemOnc.org (no financial compensation, unrelated to this work); Funding from American Association for Cancer Research (unrelated to this work).
Prior presentations: None.
Acknowledgements: The authors thank the Woods Foundation, the Jay Harris Junior Faculty Award, and the Joint Center for Radiation Therapy Foundation for their generous support of this work.

**Abstract**

**Purpose/Objective(s):** Radiotherapy (RT) toxicities can impair survival and quality-of-life, yet remain under-studied. Real-world evidence holds potential to improve our understanding of toxicities, but toxicity information is often only in clinic notes. We developed natural language processing (NLP) models to identify the presence and severity of esophagitis from notes of patients treated with thoracic RT.

**Materials/Methods:** Our corpus consisted of a gold-labeled dataset of 1,524 clinic notes from 124 patients with lung cancer treated with RT, manually annotated for CTCAE v5.0 esophagitis grade, and a silver-labeled dataset of 2,420 notes from 1,832 patients on whom toxicity grades had been collected as structured data during clinical care. We fine-tuned statistical and pre-trained BERT-based models for three esophagitis classification tasks: Task 1) no esophagitis vs. grade 1-3, Task 2) grade ≤1 vs. >1, and Task 3) no esophagitis vs. grade 1 vs. grade 2-3. Transferability was tested on 345 notes from patients with esophageal cancer undergoing RT.

**Results:** Fine-tuning PubmedBERT yielded the best performance. Best macro-F1 was 0.92, 0.82, and 0.74 for Task 1, 2, and 3, respectively. Selecting the most informative note sections during fine-tuning improved macro-F1 by ≥ 2% for all tasks. Silver-labeled data improved the macro-F1 by ≥ 3% across all tasks. For the esophageal cancer notes, best macro-F1 was 0.73, 0.74, and 0.65 for Task 1, 2, and 3, respectively, without additional fine-tuning.

**Conclusion:** To our knowledge, this is the first effort to automatically extract esophagitis toxicity severity according to CTCAE guidelines from clinic notes. The promising performance provides proof-of-concept for NLP-based automated detailed toxicity monitoring in expanded domains.

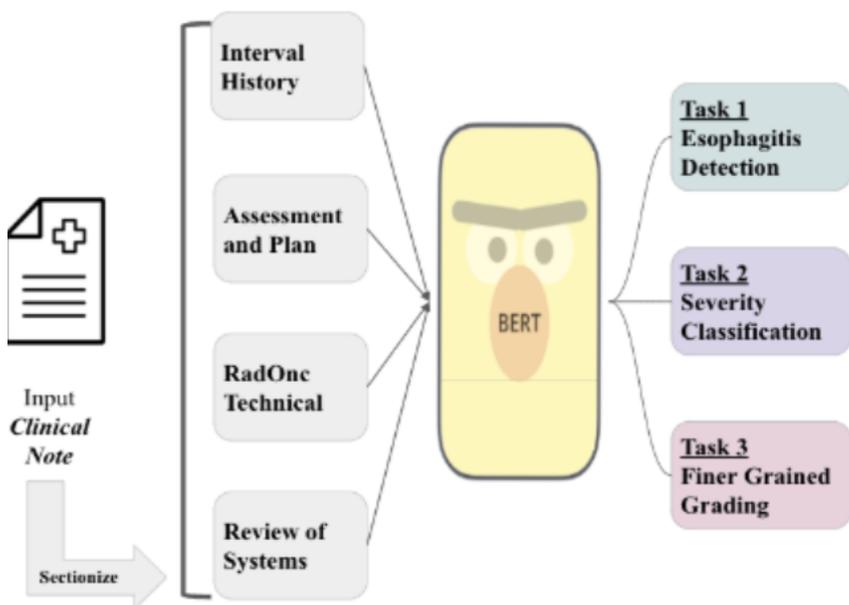

Figure 1: Clinical Pipeline

## INTRODUCTION

Cancer treatment-related adverse events (TRAEs) are an under-studied cancer outcome that have important impacts on patients' survival and quality-of-life. As cancer-specific survival improves,[1,2] a better understanding of the epidemiology, trajectories, and risk factors for TRAEs is urgently needed to support treatment decision-making and survivorship care. However, the vast majority of our understanding of TRAEs is from clinical trials, which are crucial but inherently limited in their ability to provide information on less common adverse events.[3–7]

There is enormous potential for real-world evidence (RWE) from electronic health records (EHR) to supplement adverse event information from clinical trials. Single-institutional RWE datasets have shown TRAE rates up to 10-times those reported in clinical trials.[8–13] However, a major limitation to our ability to collect high-quality RWE on adverse events is that many are almost exclusively documented in the free text of clinical notes and are not directly extractable and analyzable through conventional methods.[14–17] Natural language processing (NLP), a field of computer science that aims to convert free text into computable representations that can be used for downstream tasks, is a promising avenue to automate the extraction for adverse events documented in clinical notes. Some research exists that focuses on binary (yes/no) adverse event extraction,[18–22] however a finer-grained extraction related to the *severity* of the adverse event would further our understanding of these events.

Better methods for adverse event reporting are especially needed for radiotherapy (RT).[23] RT-related adverse events are not represented in pharmacovigilance databases, and the impact of new RT technologies and novel combinations of RT and systemic agents is under-explored.[24] In particular, lung cancer is the leading cause of death in the United States and worldwide,[1,25] and RT plays an important role in its treatment.[26–31] Esophagitis is one of the most crucial acute RT toxicities experienced by patients with lung cancer receiving thoracic RT, which inevitably exposes portions of the esophagus to radiation fields.[32] Severe esophagitis can occur in over 30% of lung cancer patients receiving RT for lung cancer,[30,32,33] and can lead to complications including malnutrition, hospitalization, and discontinuation of treatment. Importantly, esophagitis and its severity are almost exclusively documented in clinical notes.

In this study, we aimed to develop NLP methods for automated extraction of esophagitis presence and severity in the clinical notes of patients with lung cancer treated with RT according to National Cancer Institute Common Terminology Criteria for Adverse Events version 5.0 (CTCAE) guidelines.[34] Specifically, we developed NLP methods for three classification tasks: 1) none vs. grade 1-3 esophagitis, 2) grade ≤1 vs. grade 2-3 esophagitis, and 3) none vs. grade 1 vs. grade 2-3 esophagitis. No instances of grade 4 or 5 esophagitis existed in our datasets. We explored cross-domain transferability in a cohort of esophageal cancer patients.

## METHODS

### Patient population and data curation
Our study primarily used a gold-labeled lung cancer dataset of 1,524 notes from 124 non-small cell lung cancer patients who received RT at Brigham and Women's Hospital/Dana-Farber Cancer Institute between 2015-2021. We included notes written from the start of RT up to one month after the last day of RT.

| Table 1. Patient demographics and clinical characteristics | | | |
|---|---|---|---|
| Patients | Lung Cancer Dataset (n=124) | Silver-Labeled Dataset (n=1832) | Esophageal Cancer Dataset (n=75) |
| **Age in Years (Mean, SD)** | 68.6, 12.1 | 67.9, 14.6 | 70.8, 10.9 |
| **Gender** | | | |
| Male | 53 (42.7%) | 932 (50.8%) | 62 (82.7%) |
| Female | 71 (57.3%) | 900 (49.2%) | 13 (17.3%) |
| **Race** | | | |
| White | 113 (91.1%) | 1607 (87.7%) | 70 (93.3%) |
| Asian | 6 (4.8%) | 70 (3.8%) | 1 (1.3%) |
| Black | 0 | 75 (4.1%) | 1 (1.3%) |
| Two or more | 0 | 8 (0.4%) | 0 |
| Others | 0 | 31 (1.7%) | 3 (4.0%) |
| Not reported | 5 (4%) | 41 (2.2%) | 0 |
| **Ethnic Group** | | | |
| Non Hispanic | 118 (95.2%) | 1683 (91.8%) | 69 (92.0%) |
| Hispanic | 0 | 14 (0.8%) | 1 (1.3%) |
| Not reported | 6 (4.8%) | 135 (7.4%) | 5 (6.7%) |
| **Disease Site** | | | |
| Lung Cancer | | | |
| Non-small cell lung cancer | 124 (100%) | 798 (43.6%) | 0 |
| Other Histology | 0 | 126 (6.9%) | 0 |
| Gastrointestinal Cancer | | | |
| Esophageal Cancer | 0 | 167 (9.1%) | 75 (100%) |
| Other Histology | 0 | 129 (7.0%) | 0 |
| Others | 0 | 612 (33.4%) | 0 |
| **Concurrent Chemotherapy** | | | |
| Yes | 107 (86.3%) | NA* | 71 (94.7%) |
| No | 17 (13.7%) | NA* | 4 (5.3%) |

Abbreviations: NA = not applicable
All data presented as N (%) unless otherwise noted.
*Concurrent chemotherapy not available for the silver-labeled dataset.

Notes were manually annotated for the presence and maximum inferrable CTCAE v5.0 grade of esophagitis when the document was written. To establish consistent scoring guidelines for esophagitis, we annotated notes for the presence of a documented esophagitis event and its grade. We adhered to CTCAE v5.0 guidelines and based scoring solely on the information present within the free text of the note being annotated. Of note, pain with swallowing was considered esophagitis, but dysphagia alone was not. Thirteen percent of notes were dual-annotated by a radiation oncologist (DSB) and a trained data scientist (NKR), with an inter-rater reliability of 96%. The remaining notes were single-annotated by a data scientist (NKR). This dataset was split into a train, development, and test set in a proportion of roughly 80:10:10. Notes from the same patient appeared in only one of the sets.

We also used a silver-labeled dataset of 2,420 notes from 1,832 patients who underwent RT for any cancer diagnosis between 2015-2021, had structured esophagitis grades available in their EHR, and were absent in our gold-labeled dataset. At our institution, patients are seen by their radiation oncologist for a weekly on-treatment visit focused on managing side effects during RT. During these visits, physicians may optionally enter toxicity grade as structured data. No other providers document on treatment toxicity as structured data. Thus, this dataset only consisted of on-treatment visit notes by radiation oncologists. These structured data were taken as-is for silver labels. These were considered silver labels because they are often copy-forwarded and use variable toxicity grading rubrics over time; no additional quality control was applied to these labels. During training, we used this set of silver-labeled data for data augmentation and cross-domain purposes. In this dataset, lung cancer (927/1,832 (50.6%)), esophageal cancer (167/1,832 (9.1%)), and spine metastases (138/1,832 (7.5%)) were the most common diagnoses.

In addition, we curated a cross-domain validation set of 345 notes from 75 patients with esophageal cancer whose notes were not present in the gold or silver-labeled dataset. These notes were single-annotated by a radiation oncologist (DSB) using the same annotation guidelines as above.

Table 1 shows the patient-level characteristics of our datasets. Table 2 shows the note-level distribution of esophagitis grade in the three datasets. Of note, no instances of grade 4 or 5 esophagitis existed in any of the datasets.

| Table 2. Number of clinical notes and distribution of esophagitis grades across datasets | | | | |
|---|---|---|---|---|
| **Lung Cancer Dataset** | | | | |
| Esophagitis CTCAE grade* | Total (n=1524) | Train Set (n=1243) | Development Set (n=144) | Test Set (n=137) |
| None | 1030 (67.6%) | 869 (69.9%) | 76 (52.8%) | 85 (62.1%) |
| Grade 1 | 207 (13.6%) | 146 (11.8%) | 37 (25.7%) | 24 (17.5%) |
| Grade 2 | 187 (12.3%) | 136 (10.9%) | 27 (18.8%) | 24 (17.5%) |
| Grade 3 | 100 (6.6%) | 92 (7.4%) | 4 (2.8%) | 4 (2.9%) |
| **Silver-Labeled Dataset** | | | | |
| Esophagitis CTCAE grade* | Total (n=2420) | Train Set (n=2420) | Development Set | Test Set |
| None | 1562 (64.5%) | 1562 (64.5%) | NA | NA |
| Grade 1 | 376 (15.5%) | 376 (15.5%) | NA | NA |
| Grade 2 | 473 (19.5%) | 473 (19.5%) | NA | NA |
| Grade 3 | 9 (0.4%) | 9 (0.4%) | NA | NA |

| Esophageal Cancer Dataset | | | | |
|---|---|---|---|---|
| Esophagitis CTCAE grade* | Total (n=345) | Train Set | Development Set | Test Set (n=345) |
| None | 201 (58.3%) | NA | NA | 201 (58.3%) |
| Grade 1 | 91 (26.4%) | NA | NA | 91 (26.4%) |
| Grade 2 | 49 (14.2%) | NA | NA | 49 (14.2%) |
| Grade 3 | 4 (1.2%) | NA | NA | 4 (1.2%) |
| Abbreviations: CTCAE = National Cancer Institute Common Terminology Criteria for Adverse Events version 5.0; NA = not applicable. *No instances of grade 4-5 esophagitis. All data presented as N (%). | | | | |

This study was approved by the Mass General Brigham institutional review board, and consent was waived as this was deemed exempt human subjects research.

**Classification Task Definition**
Models were trained for three increasingly difficult tasks. In Task 1, binary models were trained to classify the presence or absence of esophagitis (none vs. grade 1-3 esophagitis). In Task 2, binary models were trained to classify the presence of clinically significant esophagitis (grade ≤1 vs. grade 2-3). This cut-off was chosen because it distinguishes patients who require any medical intervention from those who can be treated with conservative measures. In Task 3, multi-class models were trained to classify none vs. grade 1 vs. grade 2-3 esophagitis; grades 2 and 3 were combined due to the small number of observed events in each instance.

**Text pre-processing and classification models**
In addition to standard text pre-processing, we implemented a rules-based system to systematically mask and remove any templated sections, including structured toxicity scores, headers, and footers, before the text was input into the training process. Token and section distributions before and after preprocessing are shown in Supplemental Table 1.

We compared two major approaches for the three classification tasks: a statistical model as a baseline and neural models. For the statistical model, we used a term frequency-inverse document frequency (TF-IDF)[35] weighted Bag-of-Words with stochastic gradient descent (BoW+SGD).[36] For the neural models, we explored Bidirectional Encoder Representations from Transformers (BERT)-based models pre-trained on general domain text (BERT-base[37]) and on biomedical domain text (Bio-BERT v1.1,[38] ClinicalBERT,[39] PubMedBERT[40], and BioLinkBERT[41]) with fine-tuning. Models were fine-tuned with gold-labeled data alone, as well as gold and silver-labeled data.

For the BoW+SGD models, we used scikit-learn's SGDClassifier with log loss and l2 regularization, and hinge loss and 1000 estimators, respectively. This presents a baseline easily implementable method.

The BERT-based models' sequence lengths are fixed at 512 tokens. A batch size of 22 was used for different BERT-based models. Manual hyperparameter tuning revealed a learning rate of 0.00007 and a 0.2 dropout rate with weight decay for all BERT-based models leading to consistent model performance.

Given token length limitations in neural models, we provided combinations of informative note sections during BERT-based model fine-tuning.[42,43] We refined medspaCy (v0.20.2)[44] to develop a custom rule-based sectionizer. Using gold data only, we carried out ablation studies to determine the optimal combinations of the following sections for model input: Assessment and Plan, Interval History, Physical Exam, RT Technical (a section unique to some RT notes that describes the RT dose, fractionation, technique, modality, and other technical metrics), and Review of Systems. These sections were chosen as the most likely to contain informative information based on oncologist domain expertise. In notes with no Assessment and Plan or Interval History sections, the full note was used to not exclude short data-rich notes, especially non-physician notes that do not standardly have these sections. The best-performing combinations of sections were used during fine-tuning with the gold- and silver-labeled data.

**Evaluation**

During training and fine-tuning, we evaluated all models using the development set and assessed their final performance on the held-out test set. For each classification task, we report precision, recall, F1, and, for the binary tasks, area under the receiver operating characteristic and area under the precision-recall curve (Supplemental Methods). Performance is reported at the note level, which directly assesses the tasks the models were trained for. We also report results at the patient level for the best-performing models for each task, in which the maximum predicted esophagitis score in any note for a given patient was compared against the maximum predicted gold score in order to explore performance if the models were implemented to infer this common end-point for clinical trials.

Manual error analysis was conducted on the lung cancer test set.

**Comparison with structured EHR data**

To assess the completeness of esophagitis documentation in structured versus unstructured EHR data, we collected ICD-10 diagnosis codes documented after the start of RT for a narrow and broad definition of esophagitis in our lung cancer test set (Supplemental Methods). We compared the presence of an ICD-10 esophagitis diagnosis with our text-extracted results from Task 1, considering at least 1 note with an NLP-identified esophagitis event as an esophagitis diagnosis.

The code to train the neural models and process text in the pipeline is released at https://github.com/AIM-Harvard/Eso_alpha. The neural models trained on the full dataset cannot be released to protect patient privacy.

# RESULTS

**Note-level performance**

Table 3 reports the results of the best-performing BERT-based biomedical domain, BERT-based general domain, and statistical models for the lung cancer dataset. Performance was best for Task 1 (macro-F1 0.92), followed by Task 2 (macro-F1 0.82) and Task 3 (macro-F1 0.74). For all tasks, PubmedBERT-based models that were trained using both gold and silver-labeled had the best performance. The best-performing model for Task 1 and 2 included only Assessment and Plan and Interval History for notes with one of these sections; the best performing model for Task 3 included Assessment and Plan, Interval History, RT Technical, and Review of Systems for notes with either an Assessment and Plan or Interval History section.

| Table 3. Note-level model performance on the lung cancer test set. | | | | | | | |
|---|---|---|---|---|---|---|---|
| Task 1: None vs. CTCAE grade 1-3 esophagitis | | | | | | | |
| Model/Fine-Tuning Data | None F1 | Grade 1-3 F1 | Macro-F1 | Precision | Recall | AUCROC | AUPRC |
| PubMedBERT/Gold+Silver-Labeled Data | **0.93** | **0.90** | **0.92** | **0.93** | **0.92** | **0.92** | **0.90** |
| PubMedBERT/Gold Data | 0.91 | 0.86 | 0.88 | 0.89 | 0.89 | 0.89 | 0.78 |
| BERT-base/Gold+Silver-Labeled Data | 0.88 | 0.84 | 0.86 | 0.89 | 0.86 | 0.88 | 0.73 |
| BERT-base/Gold Data | 0.93 | 0.87 | 0.90 | 0.91 | 0.91 | 0.89 | 0.82 |
| BoW+SGD/Gold+Silver-Labeled Data | 0.87 | 0.77 | 0.82 | 0.83 | 0.83 | 0.82 | 0.69 |
| BoW+SGD/Gold Data | 0.89 | 0.82 | 0.86 | 0.85 | 0.85 | 0.85 | 0.74 |
| Task 2: CTCAE grade ≤1 vs. CTCAE grade 2-3 esophagitis | | | | | | | |
| Model/F1 | Grade ≤ 1 F1 | Grade 2-3 F1 | Macro-F1 | Precision | Recall | AUCROC | AUPRC |
| PubMedBERT/Gold+Silver-Labeled Data | **0.93** | **0.71** | **0.82** | **0.88** | **0.88** | **0.88** | **0.77** |
| PubMedBERT/Gold Data | 0.91 | 0.67 | 0.79 | 0.85 | 0.85 | 0.84 | 0.74 |
| BERT-base/Gold+Silver-Labeled Data | 0.88 | 0.68 | 0.78 | 0.88 | 0.83 | 0.85 | 0.75 |
| BERT-base/Gold Data | 0.90 | 0.58 | 0.74 | 0.83 | 0.84 | 0.73 | 0.43 |
| BoW+SGD/Gold+Silver-Labeled Data | 0.89 | 0.55 | 0.72 | 0.72 | 0.71 | 0.71 | 0.39 |
| BoW+SGD/Gold Data | 0.90 | 0.57 | 0.74 | 0.74 | 0.72 | 0.72 | 0.42 |
| Task 3: None vs. CTCAE grade 1 vs. CTCAE grade 2-3 esophagitis | | | | | | | |
| Model/F1 | None F1 | Grade 1 F1 | Grade 2-3 F1 | Macro-F1 | Precision | Recall | |
| PubMedBERT/Gold+Silver-Labeled Data | **0.93** | **0.60** | **0.69** | **0.74** | **0.83** | **0.84** | |
| PubMedBERT/Gold Data | 0.89 | 0.52 | 0.57 | 0.66 | 0.77 | 0.76 | |
| BERT-base/Gold+Silver-Labeled Data | 0.88 | 0.49 | 0.56 | 0.64 | 0.79 | 0.72 | |
| BERT-base/Gold Data | 0.84 | 0.46 | 0.44 | 0.58 | 0.71 | 0.70 | |
| BoW+SGD/Gold+Silver-Labeled Data | 0.88 | 0.33 | 0.55 | 0.59 | 0.59 | 0.59 | |
| BoW+SGD/Gold Data | 0.84 | 0.19 | 0.59 | 0.54 | 0.55 | 0.54 | |
| Abbreviations: CTCAE = National Cancer Institute Common Terminology Criteria for Adverse Events version 5.0; BoW+SGD = term frequency-inverse document frequency weighted Bag-of-Words with gradient boosting; AUCROC = area under the receiver operating characteristic; AUPRC = area under the precision-recall curve. | | | | | | | |
| Best performance metrics for each task are bolded. | | | | | | | |
| PubMedBERT and BERT-base models: Task 1 and 2 - Assessment and Plan and Interval History sections used for notes with one of these sections; otherwise the full note was used. Task 3 - Assessment and Plan, Interval History, RT Technical, and Review of Systems for notes with either an Assessment and Plan or Interval History section; otherwise the full note was used. | | | | | | | |
| BoW+SGD models: Full note used as input. | | | | | | | |

Table 4 reports the results of the best-performing models on the esophageal cancer dataset. For all tasks, models trained using both gold and silver-labeled data had the best performance. Best macro-F1 was lower than on the lung cancer dataset (Task 1: -0.19, Task 2: -0.083, Task 3: -0.097).

Table 4. Note-level model performance on the esophageal cancer dataset.

| Task 1: None vs. CTCAE grade 1-3 esophagitis | | | | | | | |
|---|---|---|---|---|---|---|---|
| Model/Fine-Tuning Data | None F1 | Grade 1-3 F1 | Macro-F1 | Precision | Recall | AUCROC | AUPRC |
| PubMedBERT/Gold+Silver-Labeled Data | 0.74 | **0.71** | **0.73** | **0.75** | 0.73 | **0.74** | **0.59** |
| PubMedBERT/Gold Data | **0.77** | 0.66 | 0.71 | 0.72 | 0.72 | 0.71 | 0.66 |
| BERT-base/Gold+Silver-Labeled Data | 0.67 | 0.68 | 0.68 | 0.72 | 0.68 | 0.70 | 0.55 |
| BERT-base/Gold Data | 0.73 | 0.58 | 0.66 | 0.66 | 0.67 | 0.65 | 0.53 |
| BoW+SGD/Gold+Silver-Labeled Data | 0.74 | **0.71** | **0.73** | 0.74 | **0.73** | **0.74** | **0.59** |
| BoW+SGD/Gold Data | 0.73 | 0.58 | 0.66 | 0.66 | 0.65 | 0.65 | 0.53 |
| Task 2: CTCAE grade ≤1 vs. CTCAE grade 2-3 esophagitis | | | | | | | |
| Model/F1 | Grade ≤ 1 F1 | Grade 2-3 F1 | Macro-F1 | Precision | Recall | AUCROC | AUPRC |
| PubMedBERT/Gold+Silver-Labeled Data | 0.89 | **0.58** | **0.74** | **0.88** | 0.83 | **0.79** | **0.34** |
| PubMedBERT/Gold Data | 0.91 | 0.51 | 0.71 | 0.85 | 0.84 | 0.78 | 0.33 |
| BERT-base/Gold+Silver-Labeled Data | 0.86 | 0.50 | 0.68 | 0.85 | 0.77 | 0.76 | 0.31 |
| BERT-base/Gold Data | **0.92** | 0.43 | 0.67 | 0.83 | **0.85** | 0.65 | 0.28 |
| BoW+SGD/Gold+Silver-Labeled Data | 0.83 | 0.49 | 0.66 | 0.65 | 0.76 | 0.76 | 0.31 |
| BoW+SGD/Gold Data | 0.89 | 0.46 | 0.68 | 0.66 | 0.69 | 0.69 | 0.30 |
| Task 3: None vs. CTCAE grade 1 vs. CTCAE grade 2-3 esophagitis | | | | | | | |
| Model/F1 | None F1 | Grade 1 F1 | Grade 2-3 F1 | Macro-F1 | Precision | Recall | |
| PubMedBERT/Gold+Silver-Labeled Data | **0.89** | **0.42** | **0.62** | **0.65** | **0.76** | **0.77** | |
| PubMedBERT/Gold Data | 0.80 | 0.32 | 0.48 | 0.53 | 0.66 | 0.68 | |
| BERT-base/Gold+Silver-Labeled Data | 0.74 | 0.32 | 0.55 | 0.54 | 0.64 | 0.60 | |
| BERT-base/Gold Data | 0.81 | 0.32 | 0.48 | 0.54 | 0.67 | 0.68 | |
| BoW+SGD/Gold+Silver-Labeled Data | 0.64 | 0.33 | 0.43 | 0.47 | 0.48 | 0.52 | |
| BoW+SGD/Gold Data | 0.73 | 0.23 | 0.52 | 0.49 | 0.52 | 0.54 | |

Abbreviations: CTCAE = National Cancer Institute Common Terminology Criteria for Adverse Events version 5.0; BoW+SGD = term frequency-inverse document frequency weighted Bag-of-Words with gradient boosting; AUCROC = area under the receiver operating characteristic; AUPRC = area under the precision-recall curve.
Best performance metrics for each task are bolded.
PubMedBERT and BERT-base models: Task 1 and 2 - Assessment and Plan and Interval History sections used for notes with one of these sections; otherwise the full note was used. Task 3 - Assessment and Plan, Interval History, RT Technical, and Review of Systems for notes with either an Assessment and Plan or Interval History section; otherwise the full note was used.
BoW+SGD models: Full note used as input.

**Patient-level performance**
Table 5 reports performance at the patient-level in the 13 patients in the lung cancer test set, using the best-performing models at the note level. Macro-F1 was 1.00 for Task 1, 0.92 for Task 2, and 0.49 for Task 3. Of note, for Task 3, the decrease in performance was largely driven by poor performance on the single patient with no esophagitis; all cases of grade 2-3 esophagitis were correctly identified.

Table 5. Patient-level performance on the lung cancer test set.

| Maximum esophagitis | Precision | Recall | F1 | Macro-F1 | Patient counts TP (n)/(TP+TN (n)) |
|---|---|---|---|---|---|
| Task 1: None vs. CTCAE grade 1-3 esophagitis | | | | | |
| None | 1.00 | 1.00 | 1.0 | NA | 1/1 |

Table 5. Patient-level performance on the lung cancer test set.

| | | | | | |
|---|---|---|---|---|---|
| Grade 1-3 | 1.00 | 1.00 | 1.0 | NA | 12/12 |
| Weighted overall | 1.00 | 1.00 | 1.0 | 1.0 | 13/13 |
| Task 2: CTCAE grade ≤1 vs. CTCAE grade 2-3 esophagitis | | | | | |
| Grade ≤1 | 1.00 | 0.80 | 0.89 | NA | 4/5 |
| Grade 2-3 | 0.89 | 1.00 | 0.94 | NA | 8/8 |
| Weighted overall | 0.93 | 0.92 | 0.92 | 0.92 | 12/13 |
| Task 3: None vs. CTCAE grade 1 vs. CTCAE grade 2-3 esophagitis | | | | | |
| None | 0.0 | 0.0 | 0.0 | NA | 0/1 |
| Grade 1 | 0.67 | 0.50 | 0.57 | NA | 2/4 |
| Grade 2-3 | 0.80 | 1.00 | 0.89 | NA | 8/8 |
| Weighted overall | 0.70 | 0.77 | 0.72 | 0.49 | 10/13 |

Abbreviations: CTCAE = National Cancer Institute Common Terminology Criteria for Adverse Events version 5.0; TP = true positives; TN = true negatives.

Table 6 shows patient-level performance of the 75 patients with esophageal cancer. Macro-F1 was 0.70 for Task 1, 0.69 for Task 2, and 0.58 for Task 3. For task 3, 24/26 cases of grade 2-3 esophagitis were correctly identified.

Table 6. Patient-level performance on the esophageal cancer dataset.

| Maximum esophagitis | Precision | Recall | F1 | Macro-F1 | Patient counts TP (n)/(TP+TN (n)) |
|---|---|---|---|---|---|
| Task 1: None vs. CTCAE grade 1-3 esophagitis | | | | | |
| None | 0.50 | 0.68 | 0.58 | NA | 13/19 |
| Grade 1-3 | 0.88 | 0.77 | 0.82 | NA | 43/56 |
| Weighted overall | 0.78 | 0.75 | 0.76 | 0.70 | 56/75 |
| Task 2: CTCAE grade ≤1 vs. CTCAE grade 2-3 esophagitis | | | | | |
| Grade ≤1 | 0.79 | 0.78 | 0.78 | NA | 38/49 |
| Grade 2-3 | 0.59 | 0.62 | 0.60 | NA | 16/26 |
| Weighted overall | 0.72 | 0.72 | 0.72 | 0.69 | 54/75 |
| Task 3: None vs. CTCAE grade 1 vs. CTCAE grade 2-3 esophagitis | | | | | |
| None | 0.61 | 0.58 | 0.59 | NA | 11/19 |
| Grade 1 | 0.62 | 0.33 | 0.43 | NA | 10/30 |
| Grade 2-3 | 0.59 | 0.92 | 0.72 | NA | 24/26 |
| Weighted overall | 0.61 | 0.60 | 0.57 | 0.58 | 45/75 |

Abbreviations: CTCAE = National Cancer Institute Common Terminology Criteria for Adverse Events version 5.0; TP = true positives; TN = true negatives.

**Error analysis**
Manual review of the best-performing model output revealed discrete error modes. Most commonly, false positives occurred in all models where only evidence of dysphagia was documented, but not esophagitis. Although dysphagia is, formally, a distinct diagnosis from esophagitis, they are frequently used interchangeably in practice and share overlapping signs, symptoms, and treatments. False positives were also found in negated, generic, or past descriptions of esophagitis. One patient had synchronous nausea/vomiting and esophagitis with several notes describing severe dehydration and malnutrition in which it was ambiguous, which was the driving etiology, leading to false positives.

**Comparison with structured EHR data**
The best-performing NLP model for Task 1 predicted esophagitis perfectly for all 13 patients in the test set. Of the 12 patients with esophagitis, only four were identified as having esophagitis using both the narrow- and broadly-defined set of esophagitis ICD-10 codes. Similarly, only 25/124 patients in the overall dataset had an esophagitis ICD-10 code documented.

**DISCUSSION**

This study provides proof-of-concept for the ability of NLP methods to automatically extract cancer treatment adverse event severity according to CTCAE grade from clinical notes. Our models for esophagitis extraction from the notes of patients undergoing RT for lung cancer performed well, with macro-F1 of 0.74-0.92. Unsurprisingly, performance dropped when the model was used cross-domain on notes of patients undergoing RT for esophageal cancer. We showed that our NLP methods outperformed ICD-10 codes for identifying esophagitis diagnosis, underscoring the importance of methods for text-based data extraction for adverse events. To the best of our knowledge, this is the first report of NLP methods for cancer treatment-related adverse event severity scoring, and demonstrates the potential for computational methods to support high-quality RWE generation on adverse events.

Our study builds upon a recent body of work on NLP methods for cancer-treatment related adverse events. Hong et al. (2020) used Apache clinical Text Analysis Knowledge Extraction System to extract CTCAE terms from 100 RT OTV notes, and reported an F1 of 0.25-0.91 for identified present terms, and 0.12-0.92 negated terms.[18] Gupta et al. (2021) developed patient-level classifiers of immune-related adverse events, achieving F1 scores of 0.59-0.71 for identifying the presence of the three most common immune-related adverse events.[19] A recent study of extraction of cancer symptoms—a separate but related task—reported a deep learning model that performed well at identifying 80 symptoms.[20]

Our study adds to the existing literature by providing evidence that NLP methods can extract the presence of adverse events and their severity. The impact of an adverse event on the patients' ultimate outcomes is driven by how severe it is. Adverse events, including esophagitis, can be very mild and require no intervention or treatment modifications or may progress to severe, potentially deadly symptoms requiring hospitalization.[32] While methods that extract the presence or absence of an adverse event are valuable for high-level monitoring, understanding the severity of the event is necessary for clinical decision-making and risk-stratification. In addition, severity extraction is necessary for monitoring and understanding expected trajectories of adverse events, which can guide intervention and surveillance strategies. Further, severity extraction according to a standardized grading system such as CTCAE is needed to compare trial-reported adverse event profiles to those in the real-world setting and to compare the adverse event burden across therapeutic strategies.

Our methods achieved the best results for the simplest task of identifying the presence or absence of esophagitis, not severity. The second task, which aimed to identify the presence or absence of esophagitis requiring medical intervention, required a more nuanced distinction between grade 1 and 2 esophagitis, but still performed very well for both classes. The third task aimed to identify whether a note documented no esophagitis, mild esophagitis, or esophagitis requiring medical intervention. While performance was excellent for classifying the absence of esophagitis, performance dropped for classifying grade 1 and grade 2-3 esophagitis. These findings likely relate to the subtle and somewhat subjective distinction between grade 1 and grade 2 esophagitis which may vary across providers, carry-forward errors due to documentation of previous esophagitis severity, and class imbalance. Nevertheless, our results are still solid, especially for the most clinically relevant task of identifying patients who require medical intervention for their esophagitis. Although the patient-level results are only exploratory given the small numbers, they motivate future efforts in automatically identifying maximum experienced CTCAE grades.

The cross-domain performance of our models on the notes of patients with esophageal cancer underlines the importance of domain adaptation methods. Out-of-domain performance loss is widely documented in clinical NLP [45–47] and is an active area of research.[48] While it serves as a warning when applying adverse event methods across cancer diagnoses while also suggesting the future extensibility of such methods. Simply applying the models trained on the lung cancer dataset yielded marked performance decrements, which based on manual review was largely due to tumor-induced symptoms overlapping with esophagitis symptoms in esophageal cancer patients. We also note that while no gold-labeled data esophageal cancer patient data were used to train the model, the silver-labeled dataset included 167 notes from esophageal cancer patients, so this is not a completely out-of-domain evaluation.

We took advantage of silver-labeled data to improve the performance of our models. The structured adverse event data that we used to augment our fine-tuning dataset was from a broader cancer population and were only present for a specialized note type—RT on-treatment visits—but still improved performance in the dataset including a wide variety of note and author types. The types and completeness of structured data collected during clinical encounters vary by provider, visit type, and department. Our approach may be useful for other researchers dealing with small gold-labeled datasets, who may have access to structured data for a subset of note types and wish to extract similar data from more diverse clinical encounters.

Our study has limitations that should be considered when interpreting the results. Most importantly, our datasets presented a skewed distribution with under-represented Grade 3 toxicities, and no instances of grade 4 or 5. In addition, they come from a single institution, impacting generalizability. However, the lung cancer dataset reflected notes written by various provider types from various specialties, somewhat offsetting the generalizability concern. Our patient population was primarily white, potentially limiting the generalizability of our models to patients from different population groups. We had no CTCAE grade 4-5 esophagitis events in our dataset and so we could not develop models to classify these most severe esophagitis grades. Yet, severe adverse events are more readily available as structured data, which could supplement our methods if comprehensive, structured data are available.[49] Our methods do not automatically identify what treatment is the cause of the adverse events; rather, we aimed only to extract an esophagitis event during a time window in which patients are known to be at risk for RT-related esophagitis, but most patients evaluated received concurrent chemotherapy. Accurate attribution is a focus of future work.

In conclusion, our methods provide proof-of-concept that NLP methods can extract the severity of cancer treatment-related adverse events. Such technologies may play an important role in RWE generation for risk-prediction, comparative studies, and survivorship care; Phase IV studies; and clinical trial reporting. In the future, NLP methods for adverse event extraction could support clinical care and yield improved patient quality-of-life by rapidly identifying patients who may benefit from early intervention to prevent progression.

**References**

1. Siegel, R. L., Miller, K. D., Wagle, N. S. & Jemal, A. Cancer statistics, 2023. *CA Cancer J. Clin.* **73**, 17–48 (2023).

2. Survival. https://progressreport.cancer.gov/after/survival.

3. Amery, W. K. & ISPE. Why there is a need for pharmacovigilance. *Pharmacoepidemiol. Drug Saf.* **8**, 61–64 (1999).

4. Phillips, R., Hazell, L., Sauzet, O. & Cornelius, V. Analysis and reporting of adverse events in randomised controlled trials: a review. *BMJ Open* **9**, e024537 (2019).

5. Pitrou, I., Boutron, I., Ahmad, N. & Ravaud, P. Reporting of safety results in published reports of randomized controlled trials. *Arch. Intern. Med.* **169**, 1756–1761 (2009).

6. Ioannidis, J. P. & Lau, J. Completeness of safety reporting in randomized trials: an evaluation of 7 medical areas. *JAMA* **285**, 437–443 (2001).

7. Ioannidis, J. P. A. & Lau, J. Improving safety reporting from randomised trials. *Drug Saf.* **25**, 77–84 (2002).

8. Suresh, K. *et al.* Pneumonitis in Non-Small Cell Lung Cancer Patients Receiving Immune Checkpoint Immunotherapy: Incidence and Risk Factors. *J. Thorac. Oncol.* **13**, 1930–1939 (2018).

9. So, A. C. & Board, R. E. Real-world experience with pembrolizumab toxicities in advanced melanoma patients: a single-center experience in the UK. *Melanoma Manag* **5**, MMT05 (2018).

10. Cho, J. Y. *et al.* Characteristics, incidence, and risk factors of immune checkpoint inhibitor-related pneumonitis in patients with non-small cell lung cancer. *Lung Cancer* **125**, 150–156 (2018).

11. Argnani, L. *et al.* Immune-related adverse events in the treatment of non-Hodgkin lymphoma with immune checkpoint inhibitors. *Sci. Rep.* **12**, 1753 (2022).

12. Jamieson, L. *et al.* Immunotherapy and associated immune-related adverse events at a large UK centre: a mixed methods study. *BMC Cancer* **20**, 743 (2020).

13. Sun, M. *et al.* Real-world data analysis of immune checkpoint inhibitors in stage III-IV adenocarcinoma and

squamous cell carcinoma. *BMC Cancer* **22**, 762 (2022).

14. Savova, G. K. *et al.* Use of Natural Language Processing to Extract Clinical Cancer Phenotypes from Electronic Medical Records. *Cancer Res.* **79**, 5463–5470 (2019).

15. Murphy, R. M. *et al.* Adverse drug event detection using natural language processing: A scoping review of supervised learning methods. *PLoS One* **18**, e0279842 (2023).

16. Luo, Y. *et al.* Natural Language Processing for EHR-Based Pharmacovigilance: A Structured Review. *Drug Saf.* **40**, 1075–1089 (2017).

17. Bitterman, D. S., Miller, T. A., Mak, R. H. & Savova, G. K. Clinical natural language processing for radiation oncology: A review and practical primer. *Int. J. Radiat. Oncol. Biol. Phys.* **110**, 641–655 (2021).

18. Hong, J. C., Fairchild, A. T., Tanksley, J. P., Palta, M. & Tenenbaum, J. D. Natural language processing for abstraction of cancer treatment toxicities: accuracy versus human experts. *JAMIA Open* **3**, 513–517 (2020).

19. Gupta, S., Belouali, A., Shah, N. J., Atkins, M. B. & Madhavan, S. Automated Identification of Patients With Immune-Related Adverse Events From Clinical Notes Using Word Embedding and Machine Learning. *JCO Clin Cancer Inform* **5**, 541–549 (2021).

20. Lindvall, C. *et al.* Deep Learning for Cancer Symptoms Monitoring on the Basis of Electronic Health Record Unstructured Clinical Notes. *JCO Clin Cancer Inform* **6**, e2100136 (2022).

21. Henry, S., Buchan, K., Filannino, M., Stubbs, A. & Uzuner, O. 2018 n2c2 shared task on adverse drug events and medication extraction in electronic health records. *J. Am. Med. Inform. Assoc.* **27**, 3–12 (2020).

22. Jagannatha, A., Liu, F., Liu, W. & Yu, H. Overview of the First Natural Language Processing Challenge for Extracting Medication, Indication, and Adverse Drug Events from Electronic Health Record Notes (MADE 1.0). *Drug Saf.* **42**, 99–111 (2019).

23. Smith, G. L. *et al.* Promoting the Appropriate Use of Advanced Radiation Technologies in Oncology: Summary of a National Cancer Policy Forum Workshop. *Int. J. Radiat. Oncol. Biol. Phys.* **97**, 450–461 (2017).

24. Bitterman, D. S. *et al.* Master Protocol Trial Design for Efficient and Rational Evaluation of Novel Therapeutic Oncology Devices. *J. Natl. Cancer Inst.* **112**, 229–237 (2020).

25. Sung, H. *et al.* Global Cancer Statistics 2020: GLOBOCAN Estimates of Incidence and Mortality Worldwide for 36 Cancers in 185 Countries. *CA Cancer J. Clin.* **71**, 209–249 (2021).

26. Pignon, J. P. *et al.* Lung Adjuvant Cisplatin Evaluation (LACE): A pooled analysis of five randomized clinical trials including 4,584 patients. *J. Clin. Orthod.* **24**, 7008–7008 (2006).

27. Chang, J. Y. *et al.* Stereotactic ablative radiotherapy versus lobectomy for operable stage I non-small-cell lung cancer: a pooled analysis of two randomised trials. *Lancet Oncol.* **16**, 630–637 (2015).

28. Albain, K. S. *et al.* Radiotherapy plus chemotherapy with or without surgical resection for stage III non-small-cell lung cancer: a phase III randomised controlled trial. *Lancet* **374**, 379–386 (2009).

29. Sundstrøm, S. *et al.* Hypofractionated palliative radiotherapy (17 Gy per two fractions) in advanced non-small-cell lung carcinoma is comparable to standard fractionation for symptom control and survival: a national phase III trial. *J. Clin. Oncol.* **22**, 801–810 (2004).

30. Curran, W. J., Jr *et al.* Sequential vs. concurrent chemoradiation for stage III non-small cell lung cancer: randomized phase III trial RTOG 9410. *J. Natl. Cancer Inst.* **103**, 1452–1460 (2011).

31. Antonia, S. J. *et al.* Overall survival with durvalumab after chemoradiotherapy in stage III NSCLC. *N. Engl. J. Med.* **379**, 2342–2350 (2018).

32. Baker, S. & Fairchild, A. Radiation-induced esophagitis in lung cancer. *Lung Cancer* **7**, 119–127 (2016).

33. Diao, K. *et al.* Radiation toxicity in patients with collagen vascular disease and intrathoracic malignancy treated with modern radiation techniques. *Radiother. Oncol.* **125**, 301–309 (2017).

34. *Common Terminology Criteria for Adverse Events (CTCAE)*. https://ctep.cancer.gov/protocoldevelopment/electronic_applications/docs/ctcae_v5_quick_reference_5x7.pdf (2017).

35. Rajaraman, A. & Ullman, J. D. Data Mining. in *Mining of Massive Datasets* 1–17 (Cambridge University Press, 2011).

36. Kowsari *et al.* Text Classification Algorithms: A Survey. *Information* vol. 10 150 Preprint at https://doi.org/10.3390/info10040150 (2019).

37. Devlin, J., Chang, M.-W., Lee, K. & Toutanova, K. BERT: Pre-training of Deep Bidirectional Transformers for Language Understanding. *arXiv [cs.CL]* (2018).

38. Lee, J. *et al.* BioBERT: a pre-trained biomedical language representation model for biomedical text mining. *Bioinformatics* **36**, 1234–1240 (2020).

39. Alsentzer, E. *et al.* Publicly Available Clinical BERT Embeddings. in *Proceedings of the 2nd Clinical Natural Language Processing Workshop* 72–78 (Association for Computational Linguistics, 2019).

40. Gu, Y. *et al.* Domain-Specific Language Model Pretraining for Biomedical Natural Language Processing. *arXiv [cs.CL]* (2020).

41. Yasunaga, M., Leskovec, J. & Liang, P. LinkBERT: Pretraining Language Models with Document Links. in *Proceedings of the 60th Annual Meeting of the Association for Computational Linguistics (Volume 1: Long Papers)* 8003–8016 (Association for Computational Linguistics, 2022).

42. Pomares-Quimbaya, A., Kreuzthaler, M. & Schulz, S. Current approaches to identify sections within clinical narratives from electronic health records: a systematic review. *BMC Med. Res. Methodol.* **19**, 155 (2019).

43. Rosenthal, S., Barker, K. & Liang, Z. Leveraging Medical Literature for Section Prediction in Electronic Health Records. in *Proceedings of the 2019 Conference on Empirical Methods in Natural Language Processing and the 9th International Joint Conference on Natural Language Processing (EMNLP-IJCNLP)* 4864–4873 (Association for Computational Linguistics, 2019).

44. Eyre, H. *et al.* Launching into clinical space with medspaCy: a new clinical text processing toolkit in Python. *AMIA Annu. Symp. Proc.* **2021**, 438–447 (2021).

45. Wu, S. *et al.* Negation's not solved: generalizability versus optimizability in clinical natural language processing. *PLoS One* **9**, e112774 (2014).

46. Kim, Y., Garvin, J., Heavirland, J. & Meystre, S. M. Improving heart failure information extraction by domain adaptation. *Stud. Health Technol. Inform.* **192**, 185–189 (2013).

47. Tepper, M., Capurro, D., Xia, F., Vanderwende, L. & Yetisgen-Yildiz, M. Statistical Section Segmentation in Free-Text Clinical Records. in *Lrec* 2001–2008 (2012).

48. Laparra, E., Mascio, A., Velupillai, S. & Miller, T. A Review of Recent Work in Transfer Learning and Domain Adaptation for Natural Language Processing of Electronic Health Records. *Yearb. Med. Inform.* **30**, 239–244 (2021).

49. Kalinich, M. *et al.* Prediction of severe immune-related adverse events requiring hospital admission in

patients on immune checkpoint inhibitors: study of a population level insurance claims database from the USA. *J Immunother Cancer* **9**, (2021).